# A Balanced Positional Control Architecture for a 12-DoF Quadruped Robot through Simulation-validation and Hardware Testing

Abid Shahriar

Department of Electrical and Electronic Engineering, American International University-Bangladesh, Dhaka -1229, Bangladesh

Email: abidshahriar97@gmail.com

**ABSTRACT**

A multi-joint enabled robot requires extensive mathematical calculations to determine the end-effector's position with respect to the other connective joints involved and their corresponding frames in a specific coordinate system. If a control architecture employs fewer positional constraints which cannot precisely determine the end effector's position in all quadrants of a 2D Cartesian plane then the robot is generally under-constrained, leading to challenges in accurate positioning to the end-effector across the entire plane. Consequently, only a subset of the end effector's degree of freedom (DoF) can be assigned for the robot's leg position for pose and trajectory estimation purposes. This paper introduces a novel approach and proposes an algorithm to consider a balanced control of the robot's leg position in a coordinate system so the robot's leg can be precisely determined and the DoF is not limited. Mathematical derivation of the joint angles is derived with forward and inverse kinematics, and Python-based simulation has been done to verify and simulate the robot's locomotion. Using Python-based code for serial communication with a micro-controller unit makes this approach more effective for demonstrating its application on a prototype leg its movement has been realized. The experimental prototype leg exhibits a commendable 78.9% accuracy with the simulated result, validating the robustness of our algorithm in practical scenarios. A comprehensive assessment of the control algorithm with random and continuous data point test has been conducted to ensure performance, so the algorithm can as well be deployed in a physical robot.

## 1. Introduction

Bio-inspired robots are gaining more popularity due to the capability of their ability to maneuver through uneven surfaces [1], [2] and less energy conversion is needed over wheel-based robots [3]. Kinematics analysis of quadruped robots improves the robot's precision and accurate control of the legs. Currently, real-time posture feedback and trajectory planning have been extensively studied and accurate controls of locomotion have been effectively achieved under the rapid testing and development of a control algorithm in both simulative testing and hardware testing environments.

However, quadruped robots with conventional kinematics algorithms fail to follow natural animal-like movement so they indispensably depend on complex or reduced sets of instructions and their degree of freedom is limited [4–8]. Many scholars proposed a wide range of control methods, but there are still many limitations that require to be explored in depth. In [9], the paper presents a mathematical model using the Denavit-Hartenberg algorithm to describe the kinematics of the robot, it does not mention any comparative analysis or evaluation of the model's accuracy and performance.

In [10], the authors proposed a kinematics analysis and a limb structure and its control through inverse kinematics using the Denavit-Hartenberg convention. However, a more comprehensive kinematics analysis considering an additional degree of freedom was ignored by assuming the current 2 DoF (two degrees of freedom) limb structure can hardly impact the overall mobility and adaptability of the quadruped robot. In[11], the authors present an analytic direct and inverse kinematic model and evaluate its accuracy through a static gait controller implementation, there is no extensive hardware testing or validation of it. In [12], the authors derived the kinematics equations and verified them through both simulation and mathematical results. While the Python-based simulation can provide a good estimation, the authors did not involve Micro-Controller Unit (MCU) code implementation to check real leg interaction. This could have provided a more comprehensive and rigorous applicability of the kinematics. In [13], the authors experimented and proposed a new control method called the Time-Pose control method for solving inverse kinematics in quadruped robots. The authors mention that the simplified model used in the gait control layer reduces the degrees of freedom and complexity of control, but they do not provide detailed information on how this simplification is achieved. In [14,15], the authors primarily focused on kinematics and simulation validation, yet they overlooked scenarios wherein the robot's legs might be positioned in a different quadrant than the leg has been originally placed in. For instance, in climbing scenarios, the relative positioning of the robot's legs in quadrants becomes crucial for its ascent. Furthermore, in the event of a fall, recalibrating the legs to a different quadrant is imperative for the robot's stability, necessitating investigation into leg repositioning. Moreover, the absence of experiments on overall robot pose estimation using rotational matrices represents another gap. Understanding the robot's orientation in 3D space is paramount for tasks such as navigation, obstacle avoidance, and coordinated movements. Hence, research into pose estimation techniques would contribute significantly to enhancing the robot's robustness. In [4], the authors proposed an extra linkage in their designed leg as it has been claimed that it reduces energy consumption and improves static stability, however, the additional complexity in the leg design complicates the inverse kinematics calculations and could potentially affect the robot's practicality and biomimicry.

**Related Works**

Aranda [16] uses a translational matrix to calculate frame-by-frame calculation is one of the ways of dynamic analysis approach. Recently, there has been a significant focus on incorporating real-time posture feedback into trajectory planning methods. For instance, Ren et al. [17] proposed a trajectory planning method that considers the changing posture information of a quadruped robot during movement. This method involves building kinematics equations for the mobile robot and obtaining a transformation matrix between the robot's trunk posture and foot-leg joint displacement. These are important when balancing the robot. As quadruped robots can go into rough terrain, for this reason, research is going on trajectory planning and locomotion of the robots. Research from Selçuk University showed the inverse and forward kinematics of their quadruped robot [13]. A dynamic kinematics model is required to administrate dynamic stability and path tracking. For testing and developing purposes, it is an interactive way to start with simulation for visual kinematics. For simulation purposes, Mudalige et al. [18] used ROS(Robot Operating System) based Simulation Environment.

Shankhat et al. [19] researched work on kinematics and analyzed kinematics for a 2-DOF robot in Simulink and Simscape multibody to create and simulate the model of the quadruped robot. The document concludes that the two DoF leg is sufficient for the robot's movement, but an additional DOF at the hip would drastically increase the overall mobility. The author also suggests applying a controller to coordinate the angular position of all joints for the robot. To discuss an inverse and direct kinematics problem, Potts et al. [20] suggest how to find the location and orientation of the robot's body and legs, given the joint variables or vice versa, and the geometric constraints involved also the paper identifies the configurations where the robot may lose or gain a degree of freedom or where the input-output relation degenerates and how to mitigate that. Another work from Park et al. [21] solves the inverse kinematics of each leg by using a reversed-order calculation and weighted least norm solution to handle the limitation of join limit by modifying the improved Jacobian method while the algorithm shifts the COG trajectory inside the stable area formed by the supporting legs while minimizing unnecessary motion. They proposed a solution for combining Jacobian of COG (Center of Gravity) and centroid of the support polygon with foot contact constraints. Zhang et al. [22] simulate the walking process of the robot in ADAMS software. Oak et al. [4] introduced a bar mechanism type of robotic leg design and linkage system and derived the forward and inverse kinematics using the D-H algorithm. Yan et al. [5] approach to full-body kinematics analysis is more dynamic and versatile as they used frame-to-frame conversion for the kinematics calculation. They also consider three different topologies for the robot: standing, walking, and floating, and derive the corresponding constraint matrices and dynamic equations. Morlando et al. [6] introduced a whole-body controller for quadruped robots that can reject external disturbances using a momentum-based observer. They implemented a motion planner in ROS (Robot Operating System) and observed the performance. Farshidian et al. [7] propose a feedback control structure that increases the robustness of the model uncertainties, actuator dynamics, delays, contact surface stiffness, and unobserved ground profiles. Chen et al. [8] approach to the kinematics solution which involves solving a sixteenth-degree polynomial of a single variable.

According to the research summarized, various approaches to kinematics analysis and control in quadruped robots have been proposed. While some studies focus on analytical models and simulation validation, others overlook crucial aspects such as additional degrees of freedom in limb structures and real leg interaction through microcontroller unit (MCU) implementation. Moreover, simplification techniques in gait control methods are often mentioned without detailed explanations. Ignoring scenarios where legs may be positioned differently and the absence of experiments on overall robot pose estimation represent further gaps. Additionally, proposals to enhance stability through leg design alterations raise concerns about increased complexity impacting practicality.

Based on the quadruped robot with 12 Degrees of Freedom (DoFs), a set of algorithms is proposed in this paper addressing the current deficiencies in kinematics analysis. Direct and Inverse kinematics equations have been derived and mapped with coordinates that the leg stays in any desired bunds and the algorithm can output correct calculations and positions generating precise angles, therefore integrating a mapping algorithm. The rotational matrix theory summarized in this paper is utilized to adjust roll, pitch, and yaw angles while pose estimating. This approach overcomes the limitations of simulation-based tests and offers a more effective solution for practical applications.

While the existing simulation model and approach to kinematics solving use rigorous calculation, our study in this article reduces the redundancy of calculation and implements a simple model. In this article, we propose a model to illustrate the kinematics of the robot which has the advantages of swiftly modifying code and provides an easy way to simulate. Since our approach to solving kinematics is Python-based, using it to solve kinematics allowed for simple communication with Arduino (MCU) using serial communication and this approach has predominance on top of only simulation-based tests and therefore an effective approach to the solution. A control architecture is proposed incorporating simulation and communicating to MCU (Micro-controller unit). The whole controlling process is divided into three phases. Utilizing the control algorithm, the estimated pose is realized in the simulative environment with full-body kinematics. The MCU sends the received computed angles of a single leg with rotation as signals to the servos to control the leg movements. Feedback is received from the leg's hardware and printed on the computing terminal.

The contribution of the paper to the journal are as follows:

1. Addressed a detailed kinematics analysis and derived equation of a 12-DoF Quadruped robot's leg and mapping of the leg's position in quadrant.
2. Proposed a control architecture for precise positional control of a quadruped robot.
3. Demonstrated a Real-time full-body kinematics simulation of the robot's pose estimation with additional parameters using the rotational matrix, validation, and trajectory planning
4. Hardware experimental validations have been conducted. Prototype leg interaction and validation based on a control algorithm developed for simulation to MCU code deployment.

This paper is organized into the following sections to address. In section 2, the robot's model is involved and analyzed, and the process for solving the kinematics is illustrated. A control architecture is elucidated with the algorithm involved. In section 3, the simulating experiment is done in a Python environment to validate the feasibility of the proposed algorithm, and the trajectory of the algorithm is estimated. Hardware-based experiment and results is also provided for the validation of the algorithm's deployment to a real robot. In section 4, the discussion of this study, limitations, and scope of future work has been suggested. In section 5, the overall research work of the simulation, control algorithm and hardware testing has been summarized.

## 2. Methods

This section will introduce the necessary dependencies to develop a control architecture through dynamic analysis. The node chain of the architecture is depicted in **Figure 5**. The approach in methods of this section is divided into 4 criteria: Forward kinematics analysis, Inverse kinematics analysis, Rotational matrix implementation, and Control architecture deployment. This section will provide extensive mathematical modelling, planning and calculations required for the control architecture development.

### 2.1. Forward kinematics analysis

An elementary step of the mathematical calculation of a multi-joint enabled robot depends on forward kinematics calculation. The forward kinematics takes the positional data in x, y, and z coordinates and generates angles of the locomotive joints to move into that specific location of the system coordinate. We created a free-body diagram in **Figure 1** to depict the simplified quadruped robot model for frame-by-frame calculation.

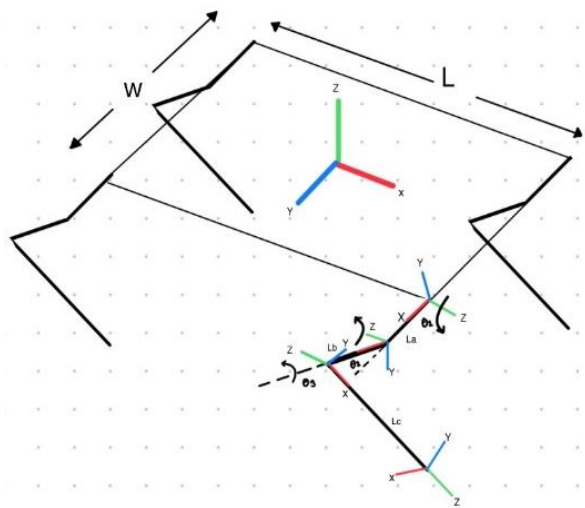

**Fig. 1.** Simplified quadruped robot free-body model for frame-by-frame calculation

In our kinematics analysis, we intend to include the Denavit-Hartenberg (DH) parameters, translational matrix, and rotational matrix for kinematics calculation. The utilization of the rotational matrix facilitates the manipulation of the robot's body orientation, including roll, yaw, and pitch, providing efficient control and manipulation techniques. The simplified quadruped robot model depicted in **Figure 1** is utilized for frame-by-frame calculations. By combining the rotation and translation matrices, it becomes feasible to construct a homogeneous transformation matrix for each component. The computation of individual transformation matrices, involving three rotations and three translations within each frame of the system, can be achieved using **equation 3.1**.

$$M_{f-1}^{f} = M_z(\alpha)M_y(\beta)M_z(\gamma) = \begin{bmatrix} M_{11} & M_{12} & M_{13} & M_{14} \\ M_{21} & M_{22} & M_{23} & M_{24} \\ M_{31} & M_{32} & M_{33} & M_{34} \\ M_{41} & M_{42} & M_{43} & M_{44} \end{bmatrix} \quad (3.1)$$

Using **equation 3.1** and the D-H parameters of **Table 1**, it is possible to create a transformation matrix for each frame of the leg of the robot. Starting from the first transformation matrix, from the first joint there is a rotation of 90° along the y-axis, therefore the equation yields,

$$T_0^1 = \begin{bmatrix} 0 & 0 & 1 & 0 \\ 0 & 1 & 0 & 0 \\ -1 & 0 & 0 & 0 \\ 0 & 0 & 0 & 1 \end{bmatrix} \quad (3.2)$$

In the next translation, there is a rotation of 90° along the x-axis which translates to length La corresponding to the coxa to femur joint leg, La, as shown in **Figure 1**, in-direction to the x-axis, and a rotation of $\theta_1$ along the z-axis, therefore the translation matrix yields,

$$T_1^2 = \begin{bmatrix} cos(\theta_1) & 0 & sin(\theta_1) & Lacos(\theta_1) \\ sin(\theta_1) & 1 & -cos(\theta_1) & Lasin(\theta_1) \\ 0 & 1 & 0 & 0 \\ 0 & 0 & 0 & 1 \end{bmatrix} \quad (3.3)$$

The Third transformation matrix, there is a rotation along the z-axis and a translation of the leg length Lb, as shown in **Figure 1**, which is Femur to Tibia connecting Joint Leg length, and the equation yields,

$$T_2^3 = \begin{bmatrix} cos(\theta_2) & -sin(\theta_2) & 0 & Lbcos(\theta_2) \\ sin(\theta_2) & cos(\theta_2) & 0 & Lbsin(\theta_2) \\ 0 & 0 & 1 & 0 \\ 0 & 0 & 0 & 1 \end{bmatrix} \quad (3.4)$$

For the Fourth transformation matrix, there is a rotation along the z axis and a translation of length Lc along the x-axis as shown in **Figure 1**, that is Femur to End Effector Joint leg length, and the equation yields,

$$T_3^4 = \begin{bmatrix} cos(\theta_3) & -sin(\theta_3) & 0 & Lccos(\theta_3) \\ sin(\theta_3) & cos(\theta_3) & 0 & Lcsin(\theta_3) \\ 0 & 0 & 1 & 0 \\ 0 & 0 & 0 & 1 \end{bmatrix} \quad (3.5)$$

Now all these individual translational matrices can be multiplied to make a transformation matrix that defines from the leg's initial frame to the foot frame. The final matrix would obtain like an equation as,

$$T_0^4 = T_0^1 T_1^2 T_2^3 T_3^4 = \begin{bmatrix} R & S \\ 0 & 1 \end{bmatrix} \qquad (3.6)$$

In **equation 3.6**, the R represents the rotational matrix and P represents the positional matrix.

Where:

$$R = \begin{bmatrix} \sin(\theta_2 + \theta_3) & \cos(\theta_2 + \theta_3) & 0 \\ \cos(\theta_2 + \theta_3)\sin(\theta_1) & \cos(\theta_3) & -\cos(\theta_1) \\ -\cos(\theta_2 + \theta_3)\cos(\theta_1) & \sin(\theta_2 + \theta_3)\cos(\theta_1) & -\sin(\theta_1) \end{bmatrix}$$

$$S = \begin{bmatrix} Lc\sin(\theta_2 + \theta_3) + Lb\sin(\theta_2) \\ (La + Lc\cos(\theta_2 + \theta_3) + Lb\cos(\theta_2))\sin(\theta_1) \\ -\cos(\theta_1)(La + Lc\cos(\theta_2 + \theta_3) + Lb\cos(\theta_2)) \end{bmatrix}$$

Equations **3.2** to **3.6** demonstrate and define the procedure for deriving the homogeneous transformation matrices that are used to model the kinematics of the robotic leg system. Through the incorporation of the rotation and translation matrices and considering the D-H parameters as displayed in **Table 1**, the transformation matrices for each frame of the robot structure are computed. These equations support the computation of individual transformation matrices made of three rotational terms and three translation terms for each frame. As such, the resultant transformation matrices contain the spatial relationships and orientations of the frames and, hence are critical in determining the overall kinematics and motion of the leg in robot's system.

**Table 1**

DH parameters involving frames of the leg of the robot.

| | DH parameters of Quadruped Leg | | | |
|---|---|---|---|---|
| $\theta_i$ | Joint | $d_i$ | $\alpha i$ | $a i$ |
| $\theta_1$ | $T_1^2$ | 0 | 90 | La |
| $\theta_2$ | $T_2^3$ | 0 | 0 | Lb |
| $\theta_3$ | $T_3^4$ | 0 | 0 | Lc |

*2.2. Inverse kinematics analysis*

In the hardware control approach, the desired angles must be transmitted via serial communication from the processing unit to the microcontroller. The microcontroller then converts these angles into Pulse Width Modulation (PWM) signals, which are sent to the servo motor. Hence inverse kinematics plays a crucial role in locomotive robots. We have considered two views of the leg in **Table 2.** In the calculation approach, we would first consider two views of one single leg from the simplified model of legs and parameters for the calculation approach. Noticeably, as depicted in left side of **Table 2.** is the Y-Z view, which means spectating from the front view of the pseudo robot's body to which it will be implemented, and as depicted in the right the tilted X-Z view which is the side view of the simplified model of the leg.

**Table 2:**

Inverse Kinematics Calculation requirements and two views of the Freebody diagram leg.

| Mathematical Computations: Inverse Kinematics Analysis ||
|---|---|
| Known Parameters: Cartesian coordinates (x, y, z), rotational values (pitch, roll, yaw), shoulder length, and wrist length. ||
| Objective: Determining the joint angles ($\theta 1$, $\theta 2$, $\theta 3$) ||
| Configuration: Tilted X-Z View (Side View) | Configuration: Y-Z View (Front View) |
| 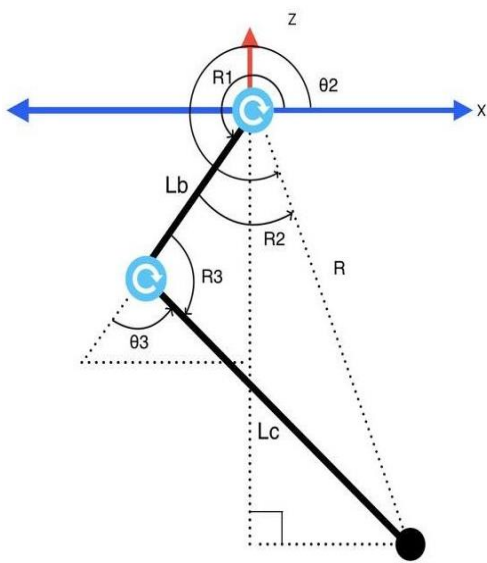 | 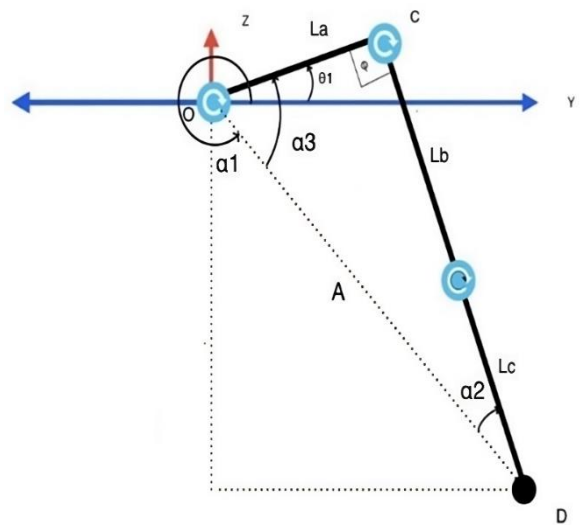 |

**Table 3**

Symbols Attributions

| Symbol | Description |
|---|---|
| La | Coxa to femur joint leg length |
| Lb | Femur to tibia joint leg length |
| Lc | Femur to end-effector joint leg length |
| $\theta_1$ | The angle (measured counterclockwise) between the positive y-axis to leg joint 'La' |
| $\theta_2$ | The angle (measured counterclockwise) between the positive y-axis to the leg end (end-effector) |
| $\theta_3$ | The angle between the femur's (Lb) extension to end -the effector |
| A | An imaginary line from rotation center to end-effector in (Y-Z) plane |
| $\alpha_1$ | The angle measured counterclockwise from the positive y-axis to line 'A' |
| $\alpha_2$ | The angle between line 'A' and 'Lc' |
| $\alpha_3$ | The angle measured counterclockwise from line 'A' to 'La' |
| $\varphi$ | The fixed angle between 'La' and 'Lb'. Design dependent, in this case, 90° |
| O, C, D | Point of rotation, connecting joint of 'La' and 'Lb', End-effector point. |
| R | An imaginary line from rotation center to end-effector in (X-Z) plane |
| R1 | The angle measured counterclockwise from the positive y-axis to line 'R' |
| R2 | The angle between 'Lb' and end-effector |
| R3 | The angle between 'Lb' and 'Lc' |
| X, Y, Z | Axis notations |
| x, y, z | Coordinate value |
| bl | body length |
| bw | body width |
| bh | body height |

Based on **Figure 2**, The first approach is to find $\alpha_1$, therefore we connect an imaginary dotted line with respect to the end-effector of the robot's leg and thus it creates a right-angled Triangle, $\triangle OCD$,

In, $\triangle OCD$, $\varphi = 90°$, A is the projection of Y-Z,

Therefore, $A = \sqrt{y^2 + z^2}$ (4.1)

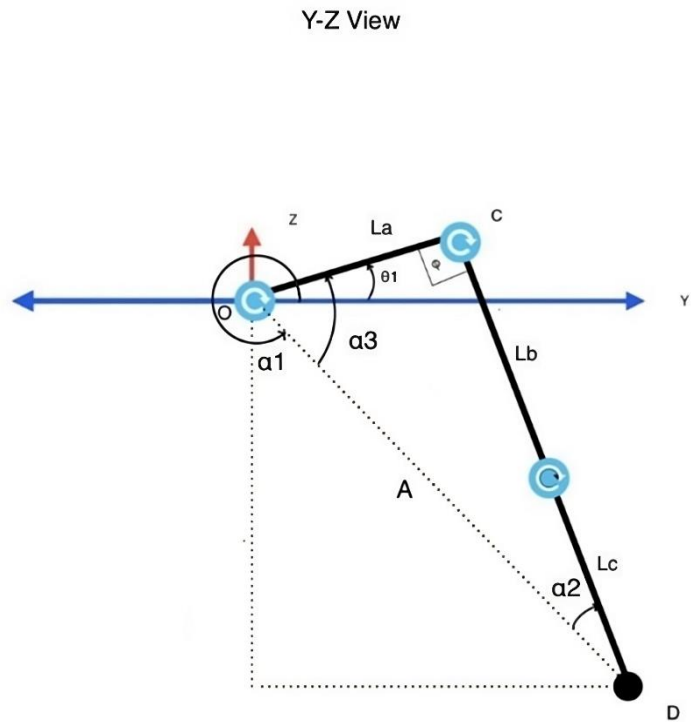

**Fig. 2.** Utilizing the front view of a single robot's leg for IK calculation to determine

Considering revolutions when measuring angles, all the rotation starts from the positive y-axis of the cartesian coordinate system. Based on the rotation of all the leg figures, La, Lb, Lc with respect to joints, the following scenario as depicted in **Figure 2**, can be the case that is from 1st to 4th co-ordinate. We have considered each case along with edge cases. For each case in our algorithm, all cases have been considered and calculated equations for $a_1$ as follows,

Case 1: Cartesian coordinate value of y and z is positive,

$a_1 = tan^{-1}\left(\frac{y}{z}\right)$ (4.2)

Case 2: The cartesian coordinate value of y is negative and z is positive,

$a_1 = -|tan^{-1}\left(\frac{y}{z}\right)| + \pi, \pi = 180°$ (4.3)

Case 3: Cartesian coordinate value of y and z is negative,

$a_1 = |tan^{-1}\left(\frac{y}{z}\right)| + \pi, \pi = 180°$ (4.4)

Case 4: The cartesian coordinate value of y is positive and z is negative,

$$a_1 = -|tan^{-1}\left(\frac{y}{z}\right)| + 2\pi, \pi = 180° \tag{4.5}$$

Case 1 and case 4 would be applicable for Right Side Legs and case 2 and case 3 is for Left side legs.

For each case, $a_1$ is varied, therefore the $\theta_1$ the the value will be varied as $\theta_1$ value can be iterated from $a_1$.

In, $\Delta OCD$,

$$\alpha_2 = sin^{-1}\left(\frac{L_a}{A} sin\Phi\right) \tag{4.6}$$

And, $\alpha_3 = (\pi - \phi - \alpha_2)$, $\phi = 90°, \pi = 180°$

or, $\alpha_3 = (90 - \alpha_2)$

or, $\alpha_3 = -sin^{-1}\left(\frac{L_a}{\sqrt{y^2+z^2}}\right) + 90°$ \qquad (4.7)

Hence, $\alpha_1 = -\alpha_3 + \theta_1$, (Considering rotation and all rotation is considered from the positive y-axis in **Figure 2**.)

Therefore, the generalized equation for $\theta_1 = \alpha_1 + \alpha_3$, [Range 0 to 2pi],

For each case with respect to the value of $\alpha_1$, $\theta_1$ value yields as follows,

While using equation (2) value for $\alpha_1$,

$$\theta_1 = tan^{-1}\left(\frac{y}{z}\right) + (90 - \alpha_2)$$

$$= tan^{-1}\left(\frac{y}{z}\right) + (90 - sin^{-1}\left(\frac{L_a}{A} sin\Phi\right))$$

$$= tan^{-1}\left(\frac{y}{z}\right) + (90 - sin^{-1}\left(\frac{L_a}{A}\right)), [\sin(90) = 1, \Phi = 90°]$$

$$= tan^{-1}\left(\frac{y}{z}\right) + (90 - sin^{-1}\left(\frac{L_a}{A}\right))$$

Or, $\theta_1 = tan^{-1}\left(\frac{y}{z}\right) + 90 - sin^{-1}\left(\frac{L_a}{\sqrt{y^2+z^2}}\right)$ \qquad (4.8)

Similarly, in other cases, $\theta_1$ value can be derived as follows,

$$\theta_1 = -|tan^{-1}\left(\frac{y}{z}\right)| + \pi - sin^{-1}\left(\frac{L_a}{\sqrt{y^2+z^2}}\right) + 90°, \text{ [equation (3) used for the value of } \alpha_1] \tag{4.9}$$

$$\theta_1 = |tan^{-1}\left(\frac{y}{z}\right)| + \pi - sin^{-1}\left(\frac{L_a}{\sqrt{y^2+z^2}}\right) + 90°, \text{ [equation (4) used for the value of } \alpha_1] \tag{4.10}$$

$$\theta_1 = -|tan^{-1}\left(\frac{y}{z}\right)| + 2\pi + 90° - sin^{-1}\left(\frac{L_a}{\sqrt{y^2+z^2}}\right), \text{ [equation (5) used for the value of } \alpha_1] \tag{4.11}$$

Since we want to limit the range of motion within 0 to $2\pi$ range because of all possible quadrants, **Algorithm 1** is used to map the value of $\theta_1$ to keep it in the desired range.

---

**ALGORITHM 1: MAPPING OF THE JOINT ANGLES AND RANGE HANDLER**

input: x, y, z
output: $\theta_1$
Result: Calculate the first joint angle, $\theta_1$ based on input coordinates y and z and mapped value from 0 to 360 range.

1  len_A = norm([0,y,z]);
2  $\alpha_1$ = point_convert_to_radian_case(y, z);
3  $\alpha_2 = \arcsin(\sin(\Phi)\frac{La}{A})$;
4  $\alpha_3 = \pi - \alpha_2 - \Phi$
5  **if** is_right **then**
6  $\quad \theta_1 = \alpha_1 - \alpha_3$;
7  **end**
8  **else**
9  $\quad \theta_1 = \alpha_1 + \alpha_3$;
10 $\quad$ **if** $\theta_1 \geq 2\pi$ **then**
11 $\quad\quad \theta_1 = \theta_1 - 2\pi$;
12 $\quad$ **end**
13 **end**

---

Next in the calculation approach, we would now consider the tilted X-Z view as it is shown in **Figure 3**, which is the side view of the simplified model of the leg.

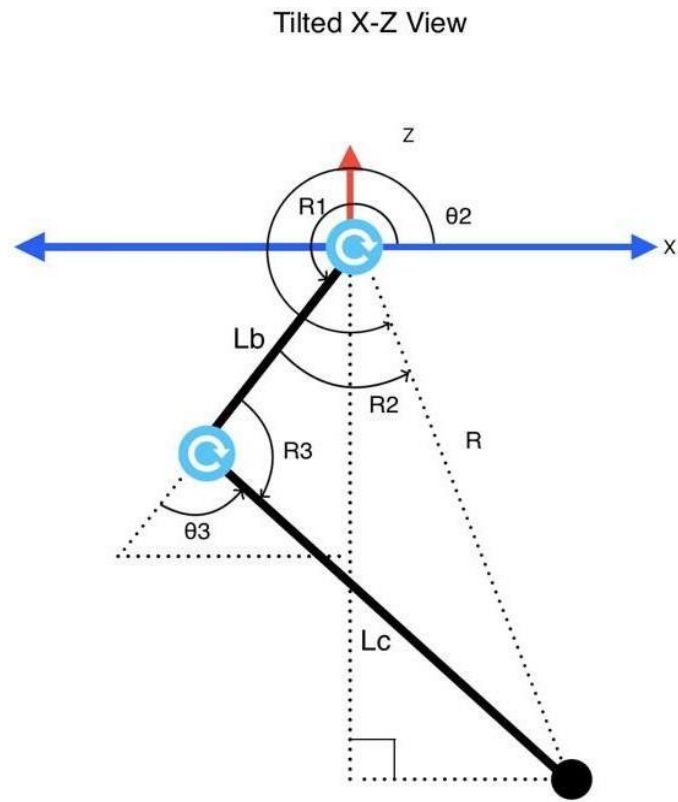

**Fig. 3.** Utilizing side-view of single robot's leg for IK calculation to determine $\theta_2$ and $\theta_3$.

From the **Figure 3**,

Deriving, the equation of R yields, $R = \sqrt{x^2 + z^2}$ (4.12)

Now, for the value of $R_1$ equation (4.13) and (4.14) is applicable.

$R_1 = tan^{-1}\left(\frac{z}{x}\right) + \pi$ (For left side legs) (4.13)

$R_1 = -\left|tan^{-1}\left(\frac{z}{x}\right)\right| + 2\pi$ (For right side legs) (4.14)

From the **Figure 3**, Applying the cosine rule yields the value of $R_2$ and $R_3$ values as,

$R_2 = cos^{-1}\left(\frac{L_b^2 + R^2 - L_c}{2.L_b.R}\right)$ (4.15)

And,

$R_3 = cos^{-1}\left(\frac{L_b^2 + L_c^2 - R^2}{2.L_b.L_c}\right)$ (4.16)

Now, eventually from the **Figure 3**,

$\theta_2 = R_1 - R_2$ (4.17)

Substituting and simplifying these terms,

Substituting R1 and R2:

$\theta_2 = \left(tan^{-1}\left(\frac{z}{x}\right) + \pi\right) - cos^{-1}\left(\frac{L_b^2 + R^2 - L_c}{2.L_b.R}\right))$ (4.18)

Substituting $R = \sqrt{x^2 + z^2}$,

$\theta_2 = \left(tan^{-1}\left(\frac{z}{x}\right) + \pi\right) - cos^{-1}\left(\frac{L_b^2 + (\sqrt{x^2+z^2})^2 - L_c}{2.L_b.R}\right)$ (4.19)

Simplifying Further,

$\theta_2 = \left(tan^{-1}\left(\frac{z}{x}\right) + \pi\right) - cos^{-1}\left(\frac{L_b^2 + x^2 + z^2 - L_c}{2.L_b.\sqrt{x^2+z^2}}\right)$ (4.20)

Eventually, now we can calculate $\theta_3$, substituting the value of $R_3$,

$\theta_3 = \pi - R_3$

$= \pi - cos^{-1}\left(\frac{L_b^2 + L_c^2 - R^2}{2.L_b.L_c}\right)$

$= \pi - cos^{-1}\left(\frac{L_b^2 + L_c^2 - (x^2 + z^2)}{2.L_b.L_c}\right)$ (4.21)

**Equation 4.12** to **4.16** demonstrate and define the procedure for deriving the robots inside the leg frame's angle through basic trigonometrical calculation. Later, we conjugate this correlation with the axis frame. Finally, with basic calculation we obtained $\theta_2$ and $\theta_3$, **equation (4.20)** and **(4.21)**, which we will be using in our control algorithm

*2.3. Implementation of Rotational Matrix*

In this section, we illustrate the need and the technique we have used to add rotational functionality to our robot's leg. For the robot to be able to have the degree of freedom individually in roll, yaw, and pitch it is therefore necessary to implement a full body rotational matrix in our algorithm to move the robot along with those rotations as shown in **Figure 4**. This is why we implemented taking rotational value as user input into the full-body kinematics. The full-body kinematics take into account to these three values and then these three values are supplied to the rotational matrix and this rotational matrix outputs as value of co-ordinates of the legs in accordance with the rotation and positioning of the leg. Finally, inverse kinematics takes these coordinates and computes the precise joint angles, enabling smooth and coordinated robot movement. This comprehensive approach grants us fine-grained control over the robot's every twist and turn. Algorithm 2 has been implemented for this process.

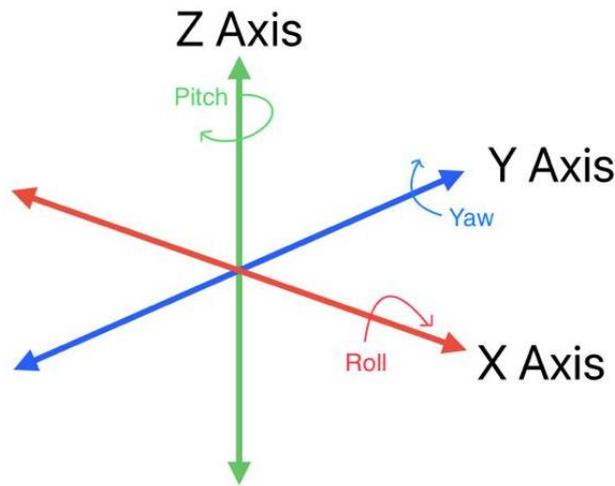

**Fig. 4.** Rotation axes of the robot

As, one of many uses of rotational algorithms in computer graphics and robotics' manipulator manipulation, we have used the common formulas **equation (5.1)**, **(5.2)**, **(5.3)** as explained in [23], by Evans (2001, pt. 3.4) in his research.

For rotation in the x-axis, y-axis, and z-axis, the matrix used respectively,

$$\mathbf{RotX} = \begin{vmatrix} 1 & 0 & 0 \\ 0 & cos(roll) & -sin(roll) \\ 0 & sin(roll) & cos(roll) \end{vmatrix} \quad , \quad (5.1)$$

$$\mathbf{RotY} = \begin{vmatrix} cos(pitch) & 0 & sin(pitch) \\ 0 & 1 & 0 \\ -sin(pitch) & 0 & cos(pitch) \end{vmatrix} \quad (5.2)$$

And,

$$\mathbf{RotZ} = \begin{vmatrix} cos(yaw) & -sin(yaw) & 0 \\ sin(yaw) & cos(yaw) & 0 \\ 0 & 0 & 1 \end{vmatrix} \quad (5.3)$$

We have used **equations (5.1)**, **(5.2)**, and **(5.3)** in **Algorithm 2**, and based on the input value we mapped the matrices to generate a 3D Rotational Matrix to work with the rotation.

| | **ALGORITHM 2: METHOD FOR CONTROLLING 3D ROTATION OF THE ROBOT'S LEG** |
|---|---|
| | input: roll, pitch, yaw |
| | output: rotationMatrix |
| | Result: Produce rotational matrix in 3D space based on input rotating angles. |
| 1 | def 3d_matrix(roll, pitch, yaw, order='xyz'): |
| 2 | rotP = np.matrix([[1, 0, 0], [0, cos(roll), -sin(roll)], [0, sin(roll), cos(roll)]]) |
| 3 | rotQ= np.matrix([[cos(pitch), 0, sin(pitch)], [0, 1, 0], [-sin(pitch), 0, cos(pitch)]]) |
| 4 | rotR = np.matrix([[cos(yaw), -sin(yaw), 0], [sin(yaw), cos(yaw), 0], [0, 0, 1] |
| 5 | rotation_orders = { |
| 6 |     'xzy': [rotY, rotZ, rotX], |
| 7 |   'xzy': [rotY, rotZ, rotX], |
| 8 |   'yxz': [rotZ, rotX, rotY], |
| 9 |   'yzx': [rotX, rotZ, rotY], |
| 10 |   'zxy': [rotY, rotX, rotZ], |
| 11 |   'zyx': [rotX, rotY, rotZ], |
| 12 |   } |
| 13 | rotartionMatrix = np.eye(3) |
| 14 | for rot in rotation_orders[order]: |
| 15 |     rotationMatrix = np.dot(rotationMatrix, rot) |
| 16 | return rotationMatrix |

### 2.4. Control architecture

The control architecture scheme, as illustrated in **Figure 5**, showcases the proposed algorithm's structure, for hardware testing and control of the robot firstly prototype leg with an Arduino micro-controller that is connected to a computer, and the communication in between is done through serial communication. In our serial communication, we have also ensured threading that way our communication protocol doesn't get overloaded. As our Algorithm goes, all our calculations are done under a class that is called the kinematics class. Under this class, there are process parameters and decision-making blocks. The robot's parameter also plays a vital role as these values are sent to the kinematics class for inverse kinematics calculation. Process parameter usually takes user input for the robot's leg end effector's position data and calculate the corresponding angles in the leg inverse kinematics calculator class. Then a data handler processes all these generated angles of all four legs and sends it for data parsing through serial communication this data is sent to Arduino MCU and if the serial is open then to the motor controller unit as PWM signals for the motors to move to the desired position.

**Table 4:**

Experiment Parameters

| La = 0.044m | bl = 0.6m |
|---|---|
| Lb = 0.28m | bw = 0.28 |
| Lc = 0.277m | bh = 0.277 |
| φ = 90° | |
| x,y,z = user input: (x, y, z) in La + Lb's bounds | |
| Roll = array ([-28, -27, -26, -25, -24……23, 24, 25, 26, 27, 28]) | |
| Yaw = array ([-21, -20, -19.-18, -17……., 18, 19, 20, 21]) | |
| Pitch = array ([-21, -20, -19.-18, -17……., 18, 19, 20, 21]) | |

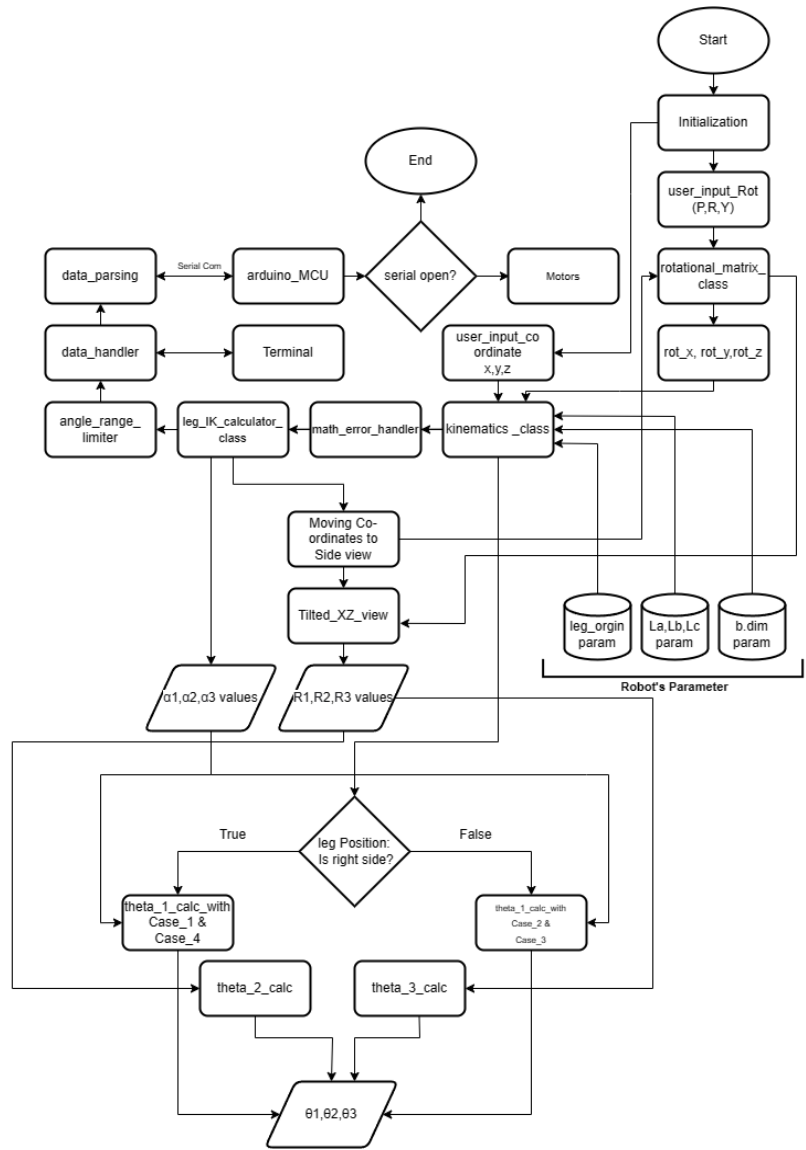

**Fig. 5.** Flow chart of the Control Architecture

The control architecture consists of different classes, each executing certain operations. The user input parameters for the coordinate and rotational degrees are processed by the user_input_coordinate and user_input_Rot parameters under kinematics_class. These parameters are processed by subclass under the math_error_handler. The math_error_handler function checks conditions, for instance, if the leg can reach a certain point and in case of an error happening when a point is too far for the leg to reach or when the math function may produce unreal value, it limits the function in the calculation so the errors could be minimized. The leg IK calculator class processes these inputs and handles necessary operations. The rotational_matrix_class manages rotating coordinates to work on the titled view of the leg and is also responsible for processing rotation while acquiring the user input for twist, tilt, and leaning and these operations are processed under two or more subclass in this method which calculates the necessary extra joints angles to calculate the desired angles. Depending on the leg position the conditional statement evaluates the correct angle for the first joint. We have implemented an angle_range_limiter function to limit the angle range to our required range as described in section 2.1. The data_handler and data_parsing functions are responsible for the output generated data for data handling and sending it to the MCU readable format to further process for the Arduino MCU and send the signal to servos.

## 3. Results

In this section, we present the outcome of our study which is divided into three main subsections. Firstly, we discuss the simulation-based experimental results, where we evaluate the stable performance of our algorithm in a virtual environment. In addition, this section delves into trajectory optimization, laying a foundational groundwork. Finally, we present a hardware-based experiment which represent the the practical validation of the method. Through these analyses, the section provides a comprehensive understanding of the capabilities of the developed control architecture.

### *3.1. Simulation-based experimental results*

In this sub-section, we have acquired the simulation results of the robot with full-body kinematics in a Python environment, where we have randomized values of the end effector's coordinate (x, y, z), and correspondingly these values were sent to inverse kinematics calculator function and therefore in generation of corresponding angles t_1, t_2, t_3 (*exact equivalent of* $\theta_1$, $\theta_2$, $\theta_3$) as illustrated in **Figure 6**. This process allowed us to assess the algorithm's ability to generate angles based on provided coordinate data. Successful angle generation was consistently achieved in the majority of cases, averaging over 50 iterations, except in instances where the coordinates reached the bounds of the two joints. To mimic the biomechanics of the simulated robot, we tested various poses in our simulation, visually analysing the stability as depicted in **Figure 6**. These positions include: (a) Stretching posture, (b) Side leg slide posture, (c) Balancing posture, (d) Standing posture (d) Crouching Posture (e) Standing tall posture. The simulation updates the spatial position of the legs in our visual environment seamlessly and the stability of the full body has been realized.

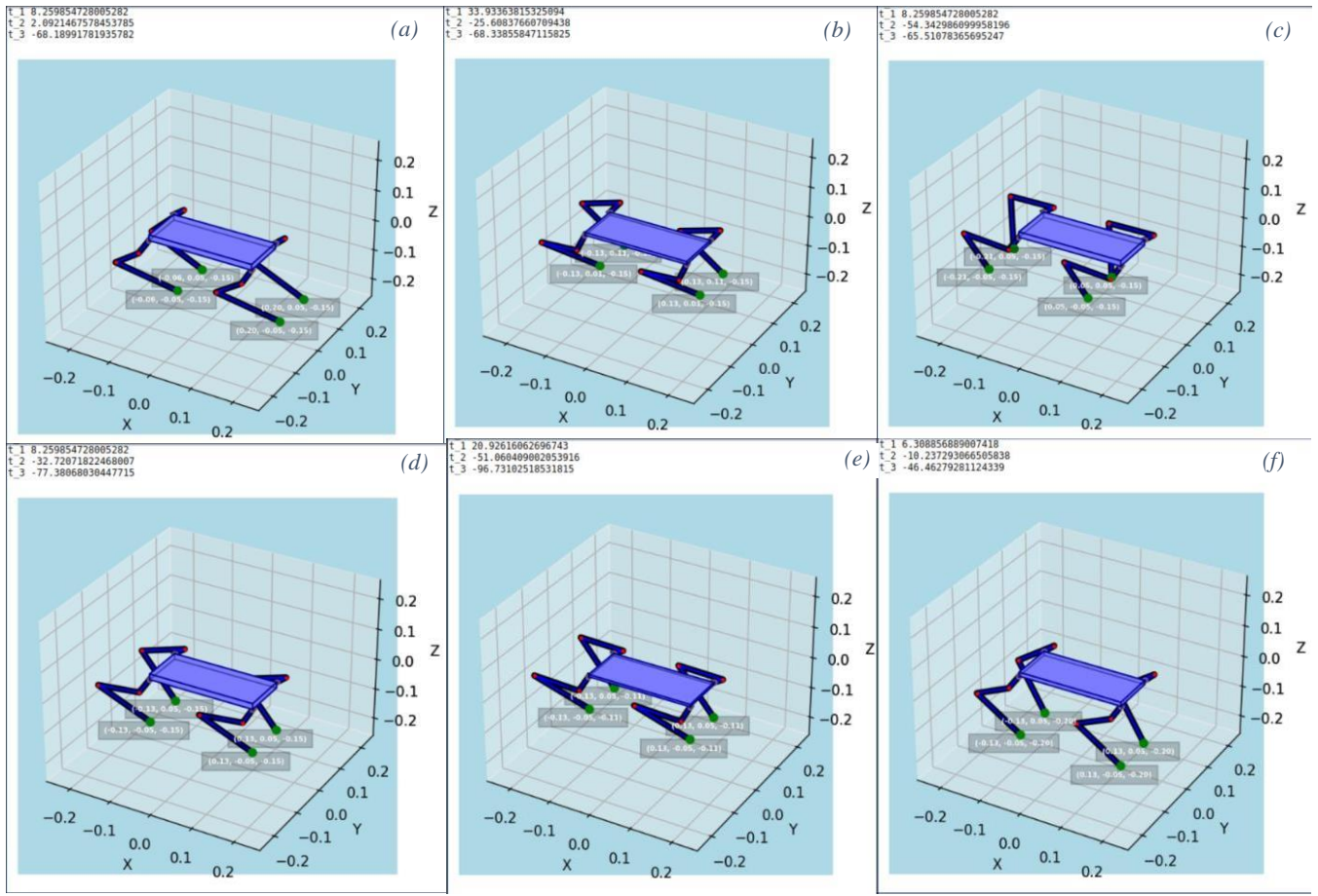

**Fig. 6.** Different pose estimation with given coordinate (x, y, z) position of the end effector and corresponding generated angles as t_1, t_2, t_3 from inverse kinematics calculation.

We have designed our control architecture to accommodate three-axis rotational input and that is used for the rotation of the body. Consequently, we individually assigned, and implemented individually assigning roll, pitch, and yaw values and visualized how they interact in the simulative environment. In our experimental result, we have found that it's interacting as expected with the given values. For the value of roll in the range of -28.0 to 28.0 degrees, pitch -21.0 to 21.0 degrees, and yaw value of 21.0 to -21.0 degrees we have tested and realized a stable posture as shown in **Figure 7**. We have achieved bio-mimicry in **Figure 7** as (a) Leaning Posture (b) Tilting Posture (c) Twisting Posture.

The findings from this section substantially demonstrate that we can utilize this algorithm with real robots for position control and mimicking different postures. The successful implementation of this algorithm with a physical quadruped robot may enhance the emulation of a wide range of postures with fidelity to natural counterparts.

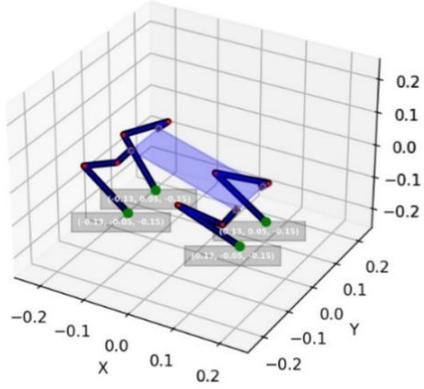
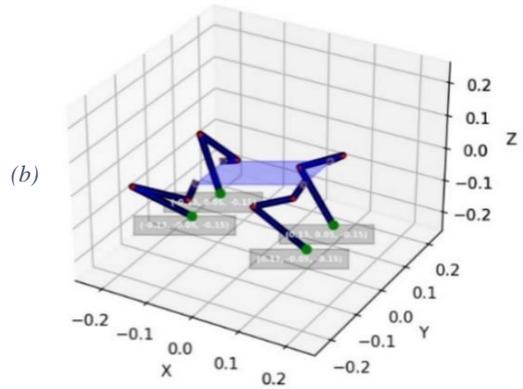
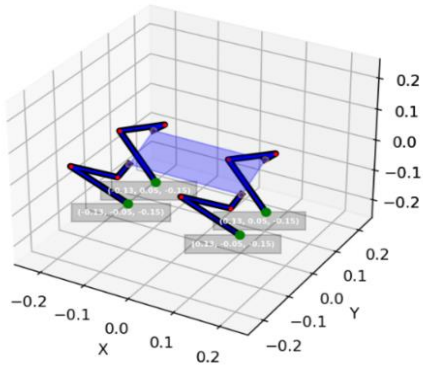
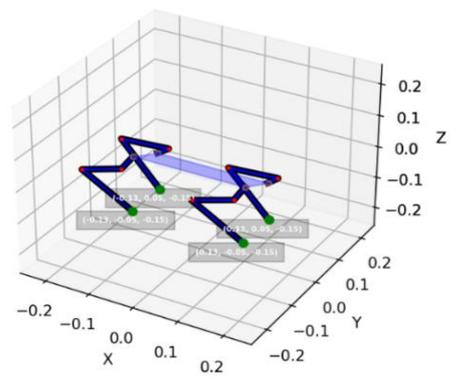
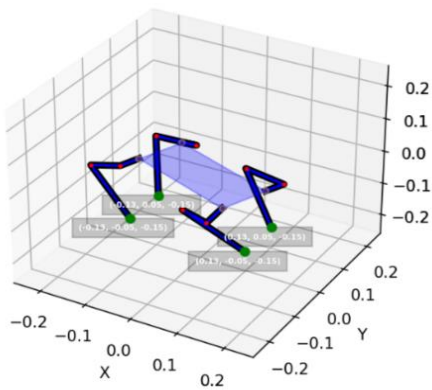
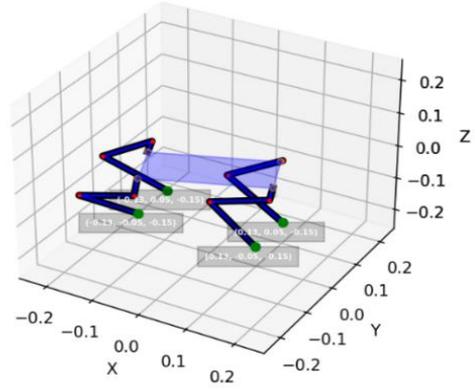

**Fig. 7.** Different pose estimation under certain roll, pitch, and yaw angles.

*3.2. Trajectory Optimization*

According research from Polet & Bertram [24] demonstrated the effectiveness of trajectory optimization in determining energetically optimal gaits in quadrupeds. This section, with **Figure 8**, illustrates the planning of the trajectory of the legs and the effectiveness of our control algorithm to follow a predefined closed-loop path. This study, in conjunction with the findings from Xi et al. [25] and Croft et al. [26], highlights the significance of trajectory planning in enhancing locomotion efficiency.

We experimented in our algorithm with a manually plotted closed-loop trajectory. The simulation resulted in **Figure 8**, a closed-loop path a curve fitted within the range of -0.2 to -0.15 unit on the x-axis and a travel distance of 0.1 unit on the y-axis. Derived from the process visual interpretation that kinematics simulation runs smooth translative motion which lays a good foundation for possible future properties and development of an interactive trot development. The trajectory path followed a closed-loop without any abruption of breakage in point which proves our algorithm's integrity to gait planning as well, leaving scope for future development. While the current trajectory path is in the shape of rectangular in our approach, the potential of defining a pre-defined path with particular coordinate points to describe a semicircle curve or cycloidal path could offer potential development in walking patterns — this factor is very well investigated through this test. To the best of our knowledge, we successfully implemented and analysed the first instance, in trajectory and basic stance of walking motion.

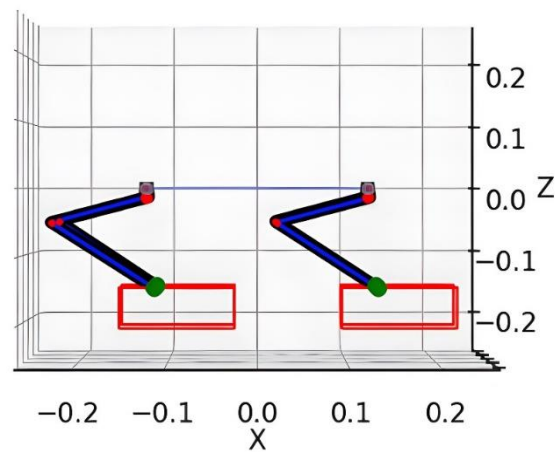

**Fig. 8.** Trajectory Estimation based on Simulation (Red rectangles indicate trajectories' bounds)

### 3.3. Hardware-based experimental results

In this approach we have tested and graded 50 points with hardware and software results and also calculated the deviation of angles. The **Figure 9**, **10**, and **11** illustrate data points which are plotted against the values of Degrees with two curves where t_1, t_2, t_3 indicate the target angle (*exact equivalent of* $\theta_1$, $\theta_2$, $\theta_3$) and t_1′, t_2′, t_3′ as the actual angle of the prototype leg. Our experiment yields 2.59% deviation for first joint, 3.64% deviation in the second joint and 20.83% deviation in third angle joint. The overall root sum square error (RSSE) is 21.10%. So, our hardware test result showed 78.9% accuracy with the software-simulated results.

It's worth noting that errors were minimized at micro step angles of 0.1 degrees, contrasting with larger errors observed at 1.0-degree intervals. These errors are influenced by several factors inherent in the capstan pulley system of the prototype leg. In the lower angle range, friction and slippage effects contribute to the smaller error observed in the experiment. It has also been noticeable that as the desired angle approaches zero, the effects of the rope elasticity are more pronounced as shown in **Figure 11**. This results in slightly elevated actual angles due to the rope stretching under tension. At higher angles, augmented friction and potential slippage in the system exacerbate errors, which leads to gradual and larger positive discrepancies.

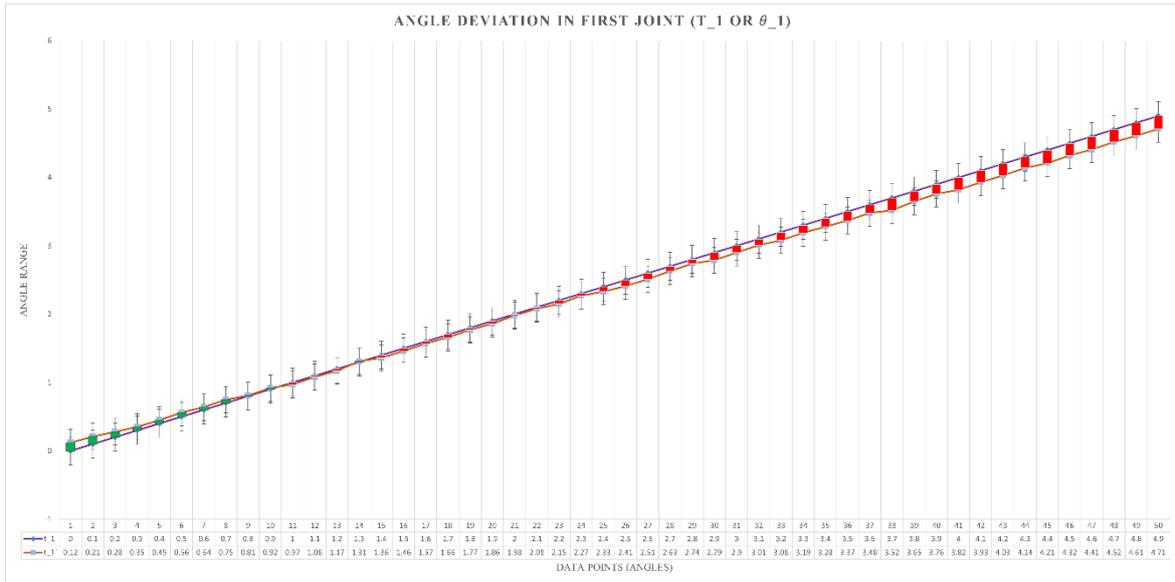

**Fig. 9.** Deviation Angle with data points of First Joint Angle, $\theta_1$

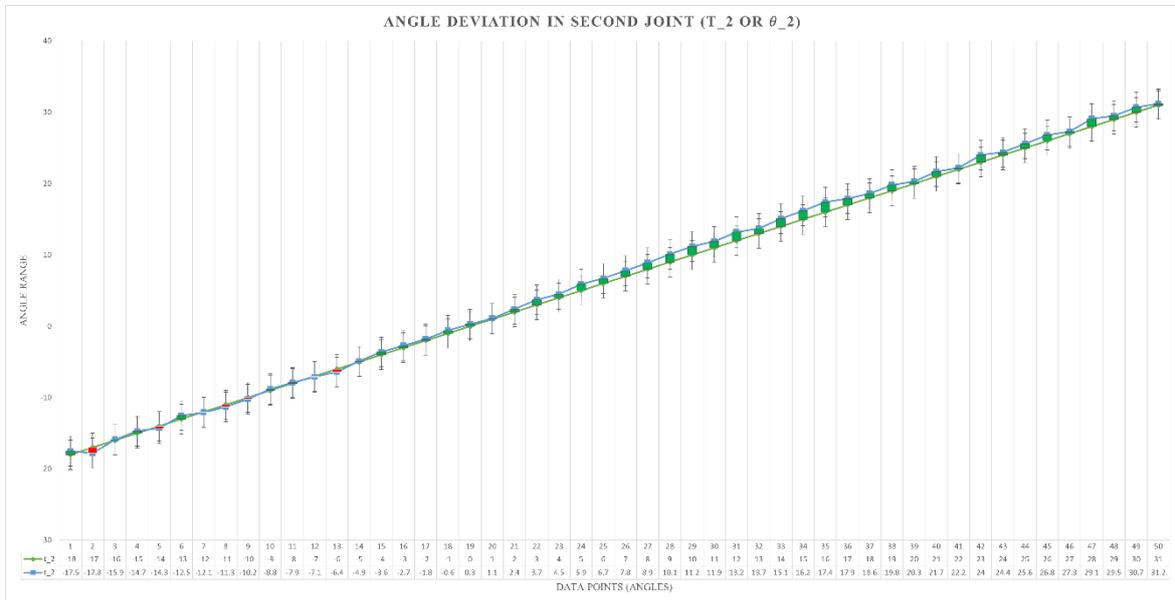

**Fig. 10.** Deviation Angle with data points of Second Joint Angle, $\theta_2$

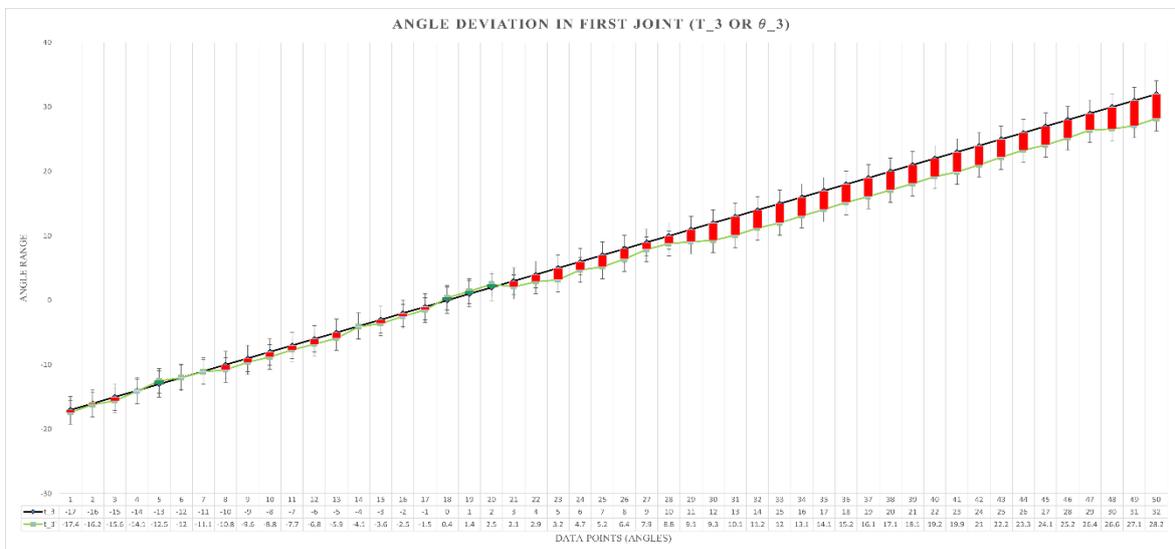

**Fig. 11.** Deviation Angle with data points of Second Joint Angle, $\theta_3$

## 4. Discussion, limitations, and future work

In this section, we discuss the findings of our study, address its limitations, and suggest directions for future research that could potentially enhance the current study and build upon the significant results obtained.

### *4.1. Discussion*:

The study presented in this paper focuses on the development and application of an algorithm for computing the full-body kinematics of a quadruped robot. Through a combination of simulation-based experiments and hardware testing with a prototype leg, the research demonstrates the successful validation of the algorithm in both virtual environments and hardware testing. To best of our knowledge, the very recent research doesn't dive too deep into validating algorithm both in a simulation environment and with hardware testing. Our research extends beyond that from kinematics analysis, and equation derivation to validation. The simulation results acquired from the testing are consistent with the mathematical calculations.

This study has several strengths. Firstly, it states a starting point for any researcher to develop a kinematics structure from scratch and derive equations for calculations. The derivation and mathematical analysis stated in this paper could be a starting point for the researcher to understand the dynamics pattern of leg movement of a bio-inspired robot and also this leaves room for further exploration and analysis with more complex leg structures and their kinematics patterns. An important part of interpreting our study is that we used 12 DoF bionic robot structures, and 3 DoF one prototype leg in our analysis. A relatively small number of recent studies were conducted with full body kinematics due to the control algorithms and hardware limitations and simulation incapability. In [16, 17, 13, 22] the authors extensively studied the dynamic movement of the leg in a simulative environment. One inherent weakness of these studies is it has been restricted validation of simulation with hardware. Also, the studies of in didn't consider implementing rotational movement of the robot body which is a crucial factor for rotative movement. We have thoroughly examined this mechanism and implemented our control algorithm, addressing the gaps related to rotative movement left by recent studies.

There has been extensive use of simulation software like MATLAB®'s SIMULINK® [29] and ADAMS® [30] for studying locomotive movements in quadruped robots and their dynamic controls. Recent studies have utilized this simulation software, but these works have been limited to simulation-based results, as they don't provide full support for hardware code deployment. This highlights the importance of our research, which contributes to both simulation and hardware implementation, addressing a significant challenge faced by many young developers in quadruped robot development. Our work leverages Python-based code for its speed and robust library support for serial communication with hardware, making this research unique in providing an open-source, code-based solution from scratch without relying on commercial software.

The simulation-based results acquired from this study have been realized in a simulative environment which exactly matches our derivation formula. In our continuous angle testing with hardware, we have acquired the result accuracy to 78.9% with hardware and software simulated results. To our utter surprise we expected a far more deviation in our hardware-based interaction with software-based interaction, since in initially we thought that the motor used in this experiment had a considerable amount of backlash and the pulley system could potentially reduce the accuracy with more amount of friction. However, the result acquired from this experiment has been satisfactory and commendable to us for prototype testing.

*4.2. Limitations*:

The proposed algorithm has been tested on only one prototype leg, although the algorithm test is done with the full body robot in the simulated environment, hence this algorithm can be used in the physical robot but the result acquired in this test cannot be fully generalized, hence we might need to analyse or use different motor parameters for further tuning to work as close to the desired angles for the deployment into the hardware. The prototype leg on which the algorithm is tested has been made with servo motors that experience backlash and these motors are not very responsive compared to BLDC (Brushless or Stepper type of motors and for these mechanical constraints our hardware test result has been varied with respect to the simulation result. Our approach in this paper has been mainly on the positional control and algorithm integrity testing which has been successfully demonstrated in our simulation result. In the hardware test, the prototype leg's interaction has been done in a certain range due to the limitation of the prototype leg's capability. The rotating body movement for pose estimation cannot be realized as it is beyond the scope of this study to realize it with one prototype leg in our hardware-based testing although the test yields a successful simulation setup. While the study showcases promising results, there are limitations to consider, such as the scalability of the algorithm to more complex robot configurations and environments.

*4.3. Future work*:

In future studies, we would like to continue the dynamic analysis and the control algorithm refinement on a more capable prototype leg build or on a full physical robot and observe the leg's interaction. Multiple process regulates dynamic movement and precise control in robots. Therefore, we propose the following future work to improve the control architecture:

1. Extend the testing of the control architecture to prototype build of the robot legs with different mechanisms and more capable motor types. The result provided in this paper suggests that the control architecture can be deployed in physical robots as our simulation test provided a commendable result. However, the testing has been limited to leg structure which is dependent on the capstan pully system and this system has substantial friction which caused a significant deviation in our hardware testing. Rigid leg structure, advanced mechanism, and uses of BLDC (Brushless Direct Current) Motor with FOC (Field Oriented Control) approach with SimpleFOC[27] can improve the algorithm's capability in hardware testing to be much more accurate and precise.
2. Consider using a more capable MCU (Micro-controller unit) for faster processing. In this work, we have used an Arduino Mega-Pro which is based on ATmega2560 and can clock speed at 16MHz [28]. For faster processing of loops in control algorithm other types of MCU can be used. STM32F4 which can be clocked at 180MHz and Arduino Due which is 84MHz are based on Cortex M4 and Cortex M3 [28]. For future development using of Cortex M based MCU can be a better deployment platform for the control system.
3. In our future work, we would like to extend this research by developing agile trot and gait patterns, building upon the basic structure implemented for the trot mechanism. We have successfully demonstrated trajectory estimation with manual input points. Depending on future needs, different trajectory patterns could be implemented, such as semi-circle, circle, cycloidal, elliptical, and parabolic paths. These advanced trajectory patterns will enhance the robot's adaptability and performance across varied terrains and operational requirements.
4. This model could benefit from state-of-the-art learning methods such as Deep Reinforcement Learning. Using this Machine learning technique allows the processing to learn how the simulation-based results differ with different hardware and leg mechanisms and structures, therefore it can reduce the error rate and improve accuracy.

## 5. Conclusions

In our study, we have extensively analyzed a 12-DoF Robot's kinematics and proposed an algorithm to tackle the precise movement of the legs. This effectively solves the pose estimation and simulation validation for the control and posture movement of a quadruped robot. This paper makes a valuable contribution to the dynamic analysis of a bio-inspired robot beyond simulation to MCU (Micro-controller unit) code deployment to hardware validation and also provides trajectory planning for future research and implications in physical robots. The algorithm test had been limited to one prototype model of the leg and due to the model leg limitation, however, with no reduced DoF of the prototype leg, we have demonstrated the hardware result in this study. Despite this limitation, the study makes important contributions because it insights into practical problems experienced by developers, makes it uniquely adequate to solve bio-mimic leg movement, and precise leg control regardless of position in quadrants which is not constrained in locomotion making the scope to utilize the proposed algorithm for self-recovery of the robot or inverted, inclined or declined movement of the robot. We tested random coordinates value to test our control architecture's integrity and tested continuous 50 random points in one iteration for the repeatability test without encountering any error. And the hardware test result showed 78.9% accuracy with the algorithm-generated results. This paper can mark a starting point for future research on solving the kinematics problem. Based on the cons of the recent state-of-the-art algorithms, new ideas can be generated to solve various gait and movement patterns, even more, DoF-enabled legged robots efficiently. Therefore, as described in this paper, the research community can start from where the literature ended.